\documentclass[conference]{IEEEtran}
\IEEEoverridecommandlockouts
\usepackage{cite}
\usepackage{amsmath,amssymb,amsfonts}
\usepackage{graphicx}
\usepackage{textcomp}
\usepackage{xcolor}
\usepackage{algpseudocode}
\usepackage{url}
\usepackage{float}
\usepackage{booktabs}
\usepackage{multirow}
\usepackage[hidelinks]{hyperref}

\def\BibTeX{{\rm B\kern-.05em{\sc i\kern-.025em b}\kern-.08em
    T\kern-.1667em\lower.7ex\hbox{E}\kern-.125emX}}

\IEEEpubid{\begin{minipage}{\textwidth}\raggedright\ \\[6pt]\noindent\rule{\textwidth}{0.4pt}\\[4pt]\centering\footnotesize
\copyright~2026 IEEE. Personal use of this material is permitted. Permission from IEEE must be obtained for all other uses, in any current or future media, including reprinting/republishing this material for advertising or promotional purposes, creating new collective works, for resale or redistribution to servers or lists, or reuse of any copyrighted component of this work in other works.
\end{minipage}}

\begin{document}

\title{SpineContextResUNet: A Computationally Efficient Residual UNet for Spine CT Segmentation\\[6pt]
{\normalfont\small\itshape 2026 IEEE 39th Int. Symp. on Computer-Based Medical Systems (CBMS), pp.\ 524--529.\ DOI: \href{https://doi.org/10.1109/CBMS69103.2026.00110}{10.1109/CBMS69103.2026.00110}}}

\author{
\IEEEauthorblockN{K S Nithurshen}
\IEEEauthorblockA{\textit{Shiv Nadar Institution of Eminence}\\
Delhi NCR, India\\
ks622@snu.edu.in}
\and
\IEEEauthorblockN{Saurabh J. Shigwan}
\IEEEauthorblockA{\textit{Shiv Nadar Institution of Eminence}\\
Delhi NCR, India\\
saurabh.shigwan@snu.edu.in}
}

\maketitle

\begin{abstract}
Automated segmentation of the vertebral column in Computed Tomography (CT) scans is a prerequisite for pathological assessment and surgical planning. However, state-of-the-art methods, particularly those based on Transformers or large-scale ensembles, demand substantial GPU resources, creating a barrier for clinical adoption in resource-constrained environments or on edge devices. To address this, we introduce \textbf{SpineContextResUNet}, a computationally efficient 3D Residual U-Net designed for rapid spinal localization. Our architecture integrates a lightweight \textit{Context Block} that employs parallel multi-dilated convolutions to capture long-range anatomical dependencies without the high latency of Recurrent Neural Networks (RNNs) or the memory overhead of Self-Attention mechanisms. Extensive validation on two public benchmarks, VerSe2020 and CTSpine1K, demonstrates that our model achieves a Dice score of 88.17\% and 88.13\% respectively. To evaluate performance under strict hardware constraints, we compared our model against a bottlenecked SwinUNETR scaled to match our $\sim$1.7M hardware footprint. While the constrained Transformer suffers severe performance degradation due to a lack of spatial inductive biases in a limited-data regime, our CNN-based approach successfully maintains high accuracy. Crucially, heavy baselines like TotalSegmentator fail due to memory exhaustion on commodity hardware (Intel Core i5, 8GB RAM), our model performs robust inference, making it a viable solution for point-of-care diagnostics and deployment on edge platforms like the Nvidia Jetson Orin Nano.

\textit{Code: https://github.com/Nithurshen/SpineContextResUNet}
\end{abstract}

\begin{IEEEkeywords}
Spine Segmentation, 3D ResUNet, Edge Computing, Context Block, Dilated Convolutions, Medical Image Analysis
\end{IEEEkeywords}

\section{Introduction}
\label{sec:introduction}

\IEEEpubidadjcol
Automated binary segmentation (localization) of the vertebral column in Computed Tomography (CT) scans is a prerequisite for computer-aided diagnosis (CADx), surgical planning, and pathological assessment of conditions such as scoliosis, vertebral fractures, and spinal stenosis. Accurate segmentation of the vertebral column allows for the rapid extraction of geometric parameters and the localization of regions of interest (ROI) for downstream tasks. However, despite the proliferation of deep learning solutions, the deployment of these systems in real-world clinical settings particularly in resource-constrained environments remains a significant challenge.

Current state-of-the-art (SOTA) approaches for spine segmentation, such as SwinUNETR \cite{he2023swinunetr} which rely on heavy Transformer-based backbones \cite{dosovitskiyimage} and nnU-Net \cite{isensee2021nnu}. While these models achieve high segmentation accuracy, they come at a prohibitive computational cost. For instance, nnU-Net requires substantial memory and high-performance computing clusters to run inference effectively. This creates a hard barrier where advanced AI diagnostic tools are inaccessible to remote hospitals in developing nations that rely on edge hardware or standard clerical workstations.

\IEEEpubidadjcol
Moreover, standard approaches often resort to slice-by-slice (2D) processing to reduce computational load. However, we argue that 2D methods are fundamentally ill-suited for spinal localization. The vertebral column is a continuous 3D anatomical structure with complex curvature (lordosis and kyphosis). In a single 2D axial slice, a vertebra can be morphologically indistinguishable from other bony structures such as the pelvic girdle or ribs. Accurate differentiation requires volumetric context along the Z-axis (craniocaudal direction), which 2D networks inherently discard. Consequently, 2D approaches often yield discontinuous segmentation masks that require heavy post-processing to stitch together.

Similarly, the recent surge of large Foundation Models, such as the Segment Anything Model (SAM) \cite{kirillov2023segment} and its medical adaptation MedSAM \cite{ma2024segment}, offers impressive zero-shot generalization but fails in this specific edge-deployment context. Foundation models are parameter-heavy. Even "Lite" versions of SAM often exceed the memory bandwidth of edge devices like the Jetson Nano, making them impractical for real-time point-of-care applications.

Furthermore, transformer-based architectures like SwinUNETR lack the spatial inductive biases inherent to CNNs. Consequently, they are data-hungry and require high parameters to effectively learn spatial hierarchies. When rigidly constrained to a low-parameter footprint (e.g., $<4$M parameters) to fit on edge devices, these Transformer models suffer substantial performance drop, particularly in medical imaging where training data is limited.

To address these limitations, we propose \textbf{SpineContextResUNet}, a lightweight model designed specifically for deployment on edge devices like the Nvidia Jetson Orin Nano. We frame the problem as a Stage 1 binary localization task, adopting a coarse-to-fine strategy \cite{zhu20183d}. In many clinical pipelines, the immediate requirement is not fine-grained identification of every vertebra (C1–L5), but rapid isolation of the spinal column as a whole. Binary spinal segmentation enables automatic region-of-interest cropping for downstream vertebra-level models, accelerates curvature and alignment analysis (e.g., scoliosis or kyphosis assessment), supports surgical navigation and trajectory planning, facilitates multi-modal image registration, and assists in trauma triage workflows. In resource-constrained environments, fast global localization can substantially reduce computational overhead by limiting subsequent processing to cropped spinal volumes rather than full-resolution CT scans.

The main contributions of this work are as follows:
\begin{itemize}
    \item We present a lightweight model, SpineContextResUNet, that achieves 88.17\% Dice on the CTSpine1K dataset and 88.13\% Dice on the VerSe2020 dataset, demonstrating robust performance across diverse data sources.
    \item We demonstrate the model's suitability for low-income healthcare settings by benchmarking inference speeds on commodity hardware (Intel Core i5, 8GB Ram), proving that SpineContextResUNet enables rapid spinal localization without the need for expensive enterprise GPUs.
\end{itemize}

\section{Proposed Method}
\label{sec:method}

The proposed architecture is engineered to balance two objectives: maximizing the receptive field to understand the global spinal structure, and minimizing the parameter count (computational cost).

\begin{figure*}[htbp]
    \centering
    \includegraphics[width=\textwidth, height=0.32\textheight, keepaspectratio, trim={0 12 0 0}, clip]{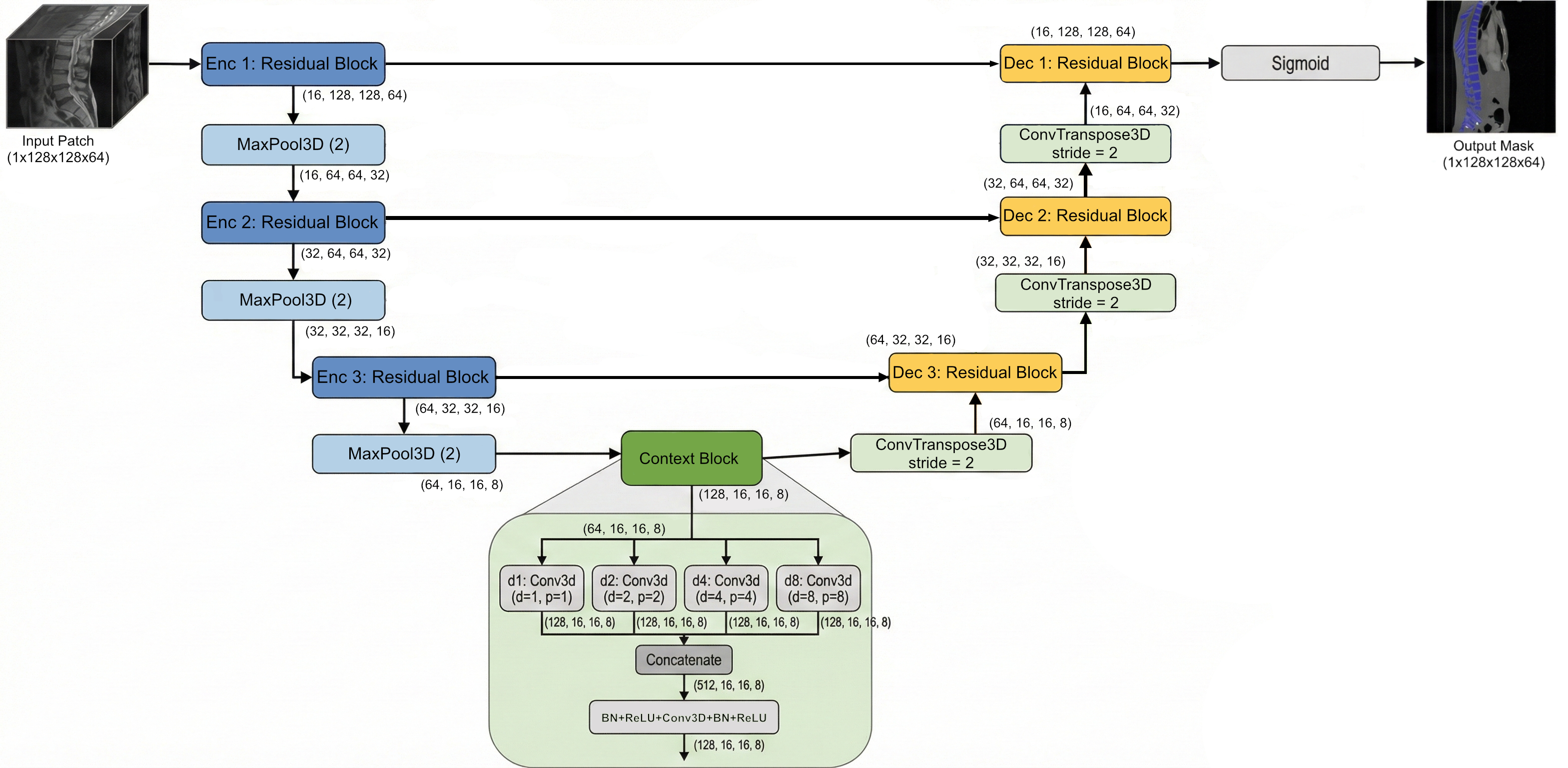}
    \caption{Overview of the proposed \textbf{SpineContextResUNet} architecture.}
    \label{fig:architecture}
\end{figure*}

\subsection{Overall Architecture}
The model follows a U-shaped encoder-decoder topology, and unlike the standard 3D U-Net \cite{cciccek20163d} which utilizes convolutional blocks, SpineContextResUNet backbone is built upon Residual Blocks \cite{diakogiannis2020resunet}. Each residual block consists of two $3 \times 3 \times 3$ convolutions with Batch Normalization (BN) and ReLU activation, linked by a residual shortcut connection. This design facilitates gradient flow and allows for the training of deeper networks without degradation.

The network processes a 3D input volume $X \in \mathbb{R}^{1 \times D \times H \times W}$. It consists of three encoder stages, a bottleneck containing the \textit{context block}, and three decoder stages. Downsampling is performed via Max Pooling ($2 \times 2 \times 2$), while upsampling utilizes trilinear interpolation followed by convolution. Skip connections concatenate high-resolution feature maps from the encoder to the decoder to recover spatial details lost during downsampling. The complete architecture is illustrated in Figure \ref{fig:architecture}.

\subsection{Context Block (ASPP)}
In spinal CT analysis, identifying the spinal column requires distinguishing it from similar bony structures (like ribs or the pelvic girdle). This requires a large effective receptive field. While previous approaches utilize Recurrent Neural Networks (RNNs) \cite{liao2018joint} or Graph Convolutional Networks (GCNs) \cite{pang2020spineparsenet} to model these global dependencies, these methods introduce significant latency.

To address this, we used a Context Block at the network bottleneck (Atrous Spatial Pyramid Pooling module \cite{chen2017deeplab}). The core logic of ASPP is to robustly segment objects at multiple scales by applying parallel atrous (dilated) convolutions. 

For a multi-dimensional input signal $x$ and a filter $w$ of size $K$, the atrous convolution yielding an output $y$ at spatial location $\mathbf{i}$ is defined as:
\begin{equation}
    y[\mathbf{i}] = \sum_{\mathbf{k}} x[\mathbf{i} + r \cdot \mathbf{k}] w[\mathbf{k}]
\end{equation}
where $r$ represents the dilation rate, determining the stride with which the input signal is sampled. Standard convolution is a special case where $r=1$. This approach effectively enlarges the field of view to capture long-range global context without downsampling the feature maps (which would sacrifice critical spatial resolution) or increasing the number of learnable parameters.

Adapted for 3D volumetric data, our Context Block aggregates this multi-scale context through four parallel branches, each employing a $3 \times 3 \times 3$ convolution with a different dilation rate $r \in \{1, 2, 4, 8\}$. Let $F_{in}$ be the input feature map to the bottleneck. The output of the branch with dilation $r$ is:
\begin{equation}
    O_r = \text{Conv}_{3\times3\times3}^{(r)}(F_{in})
\end{equation}
The branch with $r=1$ captures local texture and precise boundaries, while the branches with progressively larger rates (up to $r=8$) capture the long-range anatomical context of adjacent vertebrae. These multi-scale features are concatenated along the channel dimension and fused via a $1 \times 1 \times 1$ convolution, followed by Batch Normalization and ReLU activation which is adapted to 3D from 2D \cite{chen2017rethinking}:
\begin{equation}
    F_{out} = \sigma(\text{BN}(\text{Conv}_{1\times1\times1}(\text{Concat}[O_1, O_2, O_4, O_8])))
\end{equation}
This mechanism allows the model to process global context in a fully parallelized manner, making it significantly faster and less memory-intensive than sequential RNN-based or Transformer-based approaches.

\subsection{Loss Function}
Spinal segmentation is inherently an imbalanced problem, as the vertebrae occupy a significantly smaller percentage of the total CT volume compared to the background tissue. Relying solely on standard pixel-wise losses like Binary Cross Entropy (BCE) can lead to convergence on local minima where the model prioritizes the background. Conversely, while Dice Loss \cite{milletari2016v} effectively handles class imbalance by optimizing the overlap metric directly, it suffers from gradient instability and non-convexity, particularly during the early stages of training when predictions are highly uncertain.

To mitigate these limitations, we employ a composite loss function combining BCE and Dice Loss, a strategy validated to improve convergence stability and boundary refinement \cite{taghanaki2019combo}.
\begin{equation}
    \mathcal{L}_{Total} = \mathcal{L}_{BCE} + \mathcal{L}_{Dice}
\end{equation}

The Binary Cross-Entropy loss provides smooth gradients that facilitate learning pixel-level distributions:
\begin{equation}
    \mathcal{L}_{BCE} = - \frac{1}{N} \sum_{i=1}^{N} [y_i \log(p_i) + (1 - y_i) \log(1 - p_i)]
\end{equation}

The Dice Loss ensures the global shape and structure are preserved:
\begin{equation}
    \mathcal{L}_{Dice} = 1 - \frac{2 \sum_{i} p_i y_i + \epsilon}{\sum_{i} p_i + \sum_{i} y_i + \epsilon}
\end{equation}
where $p_i \in [0,1]$ is the predicted probability for voxel $i$, $y_i \in \{0,1\}$ is the ground truth label, and $\epsilon$ is a smoothing factor. By summing these terms, $\mathcal{L}_{BCE}$ stabilizes the training trajectory, while $\mathcal{L}_{Dice}$ drives the model to maximize the Intersection over Union (IoU), preventing the vanishing gradient problem associated with pure distribution-based losses.

\subsection{Implementation Details}
The proposed SpineContextResUNet was implemented using the PyTorch framework. Training was performed using the Adam optimizer with an initial learning rate of $1 \times 10^{-3}$ and a batch size of 4. We utilized a dynamic learning rate scheduler (\textit{ReduceLROnPlateau}) (available in PyTorch) that decays the learning rate by a factor of 10 if the validation loss plateaus for 5 epochs. To ensure spatial consistency, all input CT volumes were resampled to an isotropic resolution of $1mm^3$ and cropped or padded to a fixed patch size of $128 \times 128 \times 64$ during training.
We infer in a sliding-window approach with a stride overlap of $0.5$, and implemented a Gaussian importance weighting scheme. A Gaussian window function $W \in \mathbb{R}^{D \times H \times W}$ is computed for each patch, assigning higher importance weights to the center voxels and lower weights to the edges. Let $P_{i}$ be the predicted probability map for the $i$-th patch and $W_{i}$ be its corresponding weight window. The final reconstructed probability volume $P_{vol}$ is computed by accumulating the weighted predictions and normalizing by the weight sum:

\begin{equation}
    P_{vol} = \frac{\sum_{i} (P_{i} \cdot W_{i})}{\sum_{i} W_{i} + \epsilon}
\end{equation}

where $\epsilon$ is a small constant for numerical stability.

\section{Experiments and Results}
\label{sec:experiments}

\begin{figure*}[t]
    \centering
    \includegraphics[width=\textwidth, height=0.32\textheight, keepaspectratio]{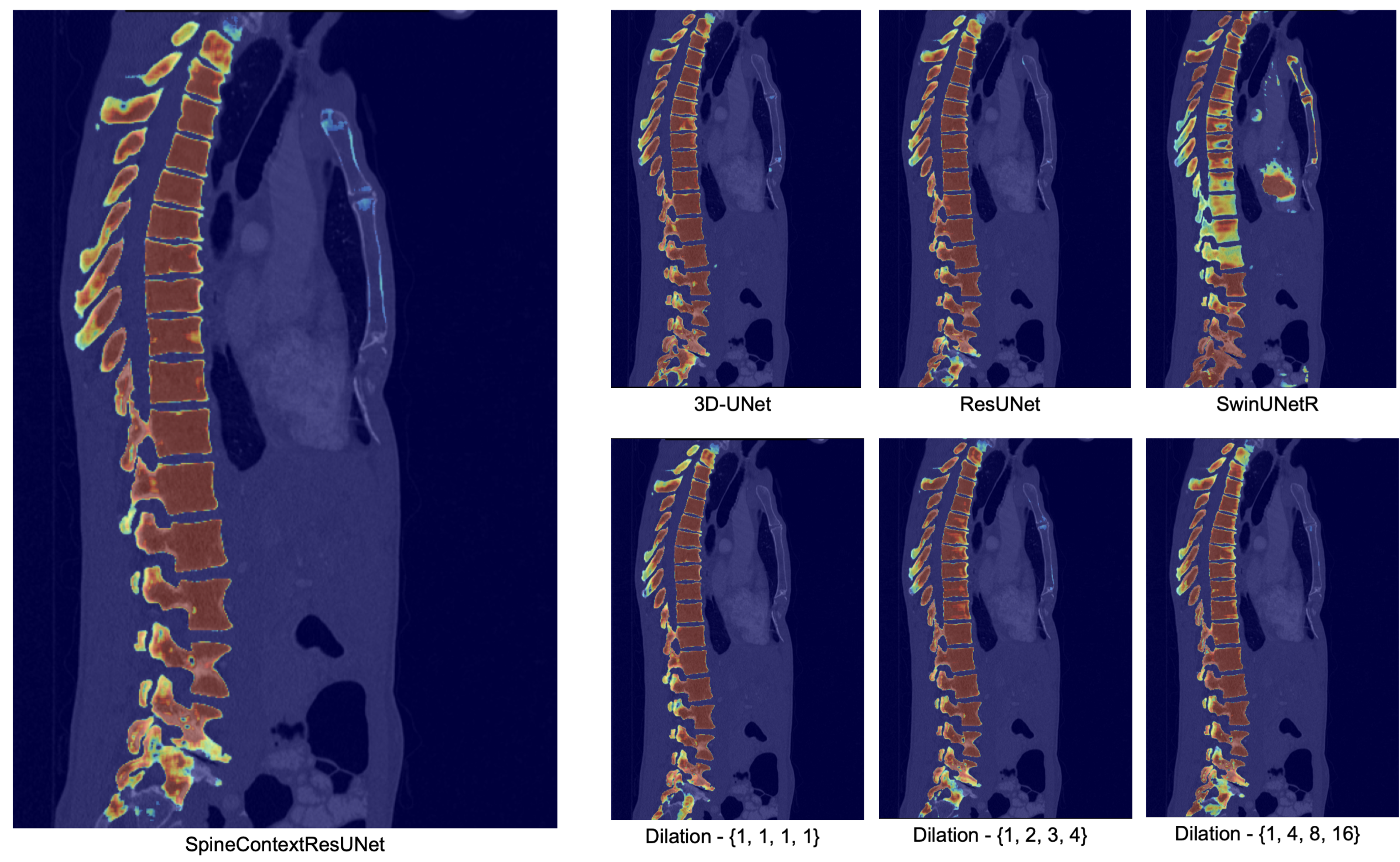}
    \caption{Grad-CAM Visualization of Vertebral Attention.}
    \label{fig:gradcam}
\end{figure*}

\subsection{Datasets and Preprocessing}
We evaluated our model on two public benchmarks: VerSe2020 (Large Scale Vertebrae Segmentation Challenge) \cite{loffler2020vertebral, sekuboyina2021verse, liebl2021computed} and CTSpine1K \cite{deng2025ctspine1k}. These datasets contain diverse spinal CTs with varying fields of view, resolutions, and pathologies, providing a robust test for generalization.
All volumes were resampled to an isotropic resolution of $1mm^3$. Intensities were clipped to the range $[-1000, 2000]$ HU and normalized to $[0, 1]$.

\subsection{Baseline Configurations}
To benchmark our model, we compared it against standard 3D U-Net \cite{cciccek20163d}, Residual UNet \cite{diakogiannis2020resunet}, and SwinUNETR \cite{he2023swinunetr}. Additionally, we included TotalSegmentator \cite{wasserthal2023totalsegmentator} as a representative baseline for the \textit{nnU-Net} framework \cite{isensee2021nnu}.

\textit{Note on TotalSegmentator:} We utilize TotalSegmentator primarily to benchmark inference latency and memory usage of state-of-the-art ensembles. TotalSegmentator is a general-purpose tool pretrained on whole-body scans. While specialized nnU-Net models trained exclusively on CTSpine1K can achieve Dice scores as high as 98.5\% \cite{deng2025ctspine1k}, they inherit the heavy computational burden of the nnU-Net self-configuring pipeline \cite{isensee2021nnu}. Our comparison highlights that SpineContextResUNet offers a trade-off: achieving acceptable Stage 1 localization accuracy (88\%) while running orders of magnitude faster on edge hardware where standard nnU-Nets fail.

\subsection{Quantitative Analysis}
Table \ref{tab:metrics} summarizes the performance on the CTSpine1K and VerSe2020 datasets.

\subsubsection{Architectural Comparison}
Our model achieves a Dice score of 88.13\% on VerSe2020 and 88.17\% on CTSpine1K, significantly outperforming the standard baselines. While the standard 3D U-Net \cite{cciccek20163d} offers the fastest inference speed (51.01s on T4 GPU), it suffers from a performance deficit, achieving only 81.32\% Dice. SpineContextResUNet trades a manageable increase in inference latency ($\approx$35s difference) for a +6.85\% improvement in Dice score. This establishes our model as an optimal middle ground, and remains deployable on edge hardware and providing improved contextual modeling compared to a standard 3D U-Net baseline. The necessity of the proposed \textit{Context Block} is evident when comparing our model against a pure ResUNet \cite{diakogiannis2020resunet}. Both architectures utilize residual connections for effective feature learning, yet the baseline ResUNet plateaus at 86.44\% Dice. This performance gap stems from the ResUNet's limited receptive field. Standard $3 \times 3 \times 3$ convolutions struggle to differentiate vertebrae from morphologically similar bony structures like ribs or the pelvic girdle. By integrating the multi-dilated Context Block, SpineContextResUNet expands its global field of view without adding significant computational overhead (only $\approx$1.5s slower inference than ResUNet on T4 GPU). This allows the model to understand the spinal column as a continuous anatomical structure rather than isolated patches. To benchmark against transformer-based architectures under strict edge-hardware limits and to ensure fair comparision, we evaluated a heavily bottlenecked SwinUNETR \cite{he2023swinunetr} to match our parameter footprint ($<4$M). As shown in Table \ref{tab:metrics}, the constrained SwinUNETR struggles significantly, yielding only 72.85\% Dice. This $\sim$15\% gap highlights that while Transformers are powerful foundation models, they are less stable under parameter constraints, limited-data regimes where they lack the inductive biases of CNNs \cite{dosovitskiyimage}.

\begin{table*}[htbp]
\renewcommand{\arraystretch}{1.15}
\caption{Segmentation Performance Comparison on VerSe2020 and CTSpine1K Datasets with Parameter Counts.}
\label{tab:metrics}
\centering
\resizebox{0.92\textwidth}{!}{%
\begin{tabular}{l|c|ccccc|ccccc}
\hline
\textbf{Architecture} & \textbf{Parameters} & \multicolumn{5}{c|}{\textbf{VerSe2020}} & \multicolumn{5}{c}{\textbf{CTSpine1K}} \\
 & & \textbf{Dice} & \textbf{IoU} & \textbf{Prec} & \textbf{Recall} & \textbf{F1} & \textbf{Dice} & \textbf{IoU} & \textbf{Prec} & \textbf{Recall} & \textbf{F1} \\ \hline
SwinUNETR \cite{he2023swinunetr} & 3,746,536 & 0.7387 & 0.6054 & 0.8904 & 0.6622 & 0.7595 & 0.7285 & 0.6031 & 0.6413 & 0.9098 & 0.7523 \\
3D U-Net \cite{cciccek20163d} & 1,788,274 & 0.8144 & 0.6949 & 0.8954 & 0.7622 & 0.8234 & 0.8132 & 0.7059 & 0.7559 & 0.8997 & 0.8216 \\
ResUNet \cite{diakogiannis2020resunet} & 1,424,545 & 0.8652 & 0.7700 & 0.9039 & 0.8407 & 0.8712 & 0.8644 & 0.7833 & 0.8479 & 0.8951 & 0.8709 \\ \hline
\textbf{SpineContextResUNet} & \textbf{1,703,841} & \textbf{0.8813} & \textbf{0.7936} & \textbf{0.9083} & \textbf{0.8617} & \textbf{0.8844} & \textbf{0.8817} & \textbf{0.8078} & \textbf{0.8838} & \textbf{0.8934} & \textbf{0.8886} \\ \hline
TotalSegmentator \cite{wasserthal2023totalsegmentator, isensee2021nnu} & 49,137,656 & 0.8748 & 0.7785 & 0.9286 & 0.8280 & 0.8754 & 0.8904 & 0.8109 & 0.8659 & 0.9310 & 0.8973 \\ \hline
\end{tabular}%
}
\end{table*}

\subsubsection{Ablation Study}
To validate the specific design of our Context Block, we compared different dilation strategies (Table \ref{tab:ablation}). The performance disparity highlights the critical relationship between the receptive field and the bottleneck feature map dimensions ($16 \times 16 \times 8$).

\begin{table*}[htbp]
\renewcommand{\arraystretch}{1.15}
\caption{Ablation Study: Impact of Different Dilation Configurations on Model Performance.}
\label{tab:ablation}
\centering
\resizebox{0.92\textwidth}{!}{%
\begin{tabular}{l|ccccc|ccccc}
\hline
\textbf{Configuration} & \multicolumn{5}{c|}{\textbf{VerSe2020}} & \multicolumn{5}{c}{\textbf{CTSpine1K}} \\
 & \textbf{Dice} & \textbf{IoU} & \textbf{Prec} & \textbf{Recall} & \textbf{F1} & \textbf{Dice} & \textbf{IoU} & \textbf{Prec} & \textbf{Recall} & \textbf{F1} \\ \hline
$\{1, 1, 1, 1\}$ & 0.8652 & 0.7700 & 0.9039 & 0.8407 & 0.8712 & 0.8644 & 0.7833 & 0.8479 & 0.8951 & 0.8709 \\
$\{1, 2, 3, 4\}$ & 0.8665 & 0.7749 & 0.9103 & 0.8456 & 0.8768 & 0.8712 & 0.7887 & 0.8521 & 0.8943 & 0.8727 \\
$\{1, 4, 8, 16\}$ & 0.8732 & 0.7844 & 0.9152 & 0.8522 & 0.8826 & 0.8756 & 0.7963 & 0.8566 & 0.8955 & 0.8756 \\ \hline
\textbf{\{1, 2, 4, 8\}} & \textbf{0.8813} & \textbf{0.7936} & \textbf{0.9083} & \textbf{0.8617} & \textbf{0.8844} & \textbf{0.8817} & \textbf{0.8078} & \textbf{0.8838} & \textbf{0.8934} & \textbf{0.8886} \\ \hline
\end{tabular}%
}
\end{table*}

\begin{table*}[htbp]
\renewcommand{\arraystretch}{1.15}
\caption{Inference Time (Seconds) across Different Hardware Configurations.}
\label{tab:inference}
\centering
\resizebox{0.92\textwidth}{!}{%
\begin{tabular}{l|c|c|c|c|c}
\hline
\textbf{Model} & \textbf{Colab CPU} & \textbf{NVidia T4} & \textbf{Apple M4 Pro CPU} & \textbf{Apple M4 Pro GPU} & \textbf{Intel Core i5 - 8500} \\
 & \textit{(Intel Xeon)} & \textit{(16GB VRAM)} & \textit{(14 Cores)} & \textit{(20 Cores)} & \textit{(8GB Ram)} \\ \hline
3D U-Net \cite{cciccek20163d} & 1856.74 & 51.01 & 199.48 & 85.91 & 348.25 \\
ResUNet \cite{diakogiannis2020resunet} & 2801.53 & 85.18 & 369.95 & 120.11 & 721.34 \\
SwinUNETR & 3166.21 & 81.39 & 293.74 & 107.26 & 735.33 \\ \hline
\textbf{SpineContextResUNet} \cite{he2023swinunetr} & \textbf{2940.15} & \textbf{86.66} & \textbf{367.98} & \textbf{124.49} & \textbf{792.49} \\ \hline
TotalSegmentator \cite{wasserthal2023totalsegmentator, isensee2021nnu} & \textit{Session Terminated} & 127.67 & 471.26 & 153.34 & \textit{Crashed} \\ \hline
\end{tabular}%
}
\end{table*}

\begin{itemize}
    \item Non-Dilated Convolutions ($\{1, 1, 1, 1\}$):
This configuration effectively mimics a standard ResUNet bottleneck. With a constant dilation of $r=1$, the receptive field is strictly limited to local neighborhoods ($3 \times 3 \times 3$). Consequently, the model struggles to distinguish the spinal column from morphologically similar bony structures, such as ribs or the pelvic girdle, resulting in the lowest Dice score (86.44\%).

    \item Linear Dilation Progression ($\{1, 2, 3, 4\}$):
While this configuration improves over the baseline by expanding the field of view, it fails to achieve optimal performance (87.12\% Dice). The linear step size results in significant overlap between the receptive fields of the $r=3$ and $r=4$ branches, leading to redundant feature extraction. Furthermore, the maximum dilation of $r=4$ is insufficient to capture the global context of the entire $16 \times 16$ feature map, leaving the model blind to long-range dependencies between distant vertebrae.

    \item Excessive Dilation ($\{1, 4, 8, 16\}$):
While increasing the receptive field is generally beneficial, we observed a performance degradation with larger rates. This is due to the sampling void phenomenon. After three downsampling stages, the spatial resolution of the feature map at the bottleneck is reduced to $16 \times 16 \times 8$ voxels. A dilation rate of $r=16$ implies a sampling stride that exceeds the physical dimensions of the feature map itself. As a result, the outermost kernels predominantly sample zero-padding (background noise) rather than valid feature information, introducing instability and gridding artifacts into the decoder.

    \item Optimality of the Proposed Configuration ($\{1, 2, 4, 8\}$):
The proposed configuration strikes the optimal balance. The varying rates capture a hierarchy of context:
        \begin{itemize}
            \item $r=1$ captures fine-grained boundaries and texture.
            \item $r=2, 4$ capture the shape of individual vertebral bodies.
            \item $r=8$ spans an effective field of $\approx 17$ voxels, which covers the entire spatial extent of the $16 \times 16$ bottleneck feature map.
        \end{itemize}
        This allows the $r=8$ branch to function as a pseudo-global attention mechanism, aggregating context from the entire visible spinal patch to ensure spatial consistency, while avoiding the padding issues seen with $r=16$.
        
\end{itemize}

\subsection{Explainability Analysis}
To validate that our model relies on relevant anatomical features rather than artifacts, we employed Gradient-weighted Class Activation Mapping (Grad-CAM) \cite{selvaraju2017grad}. We visualized the gradients flowing into the final bottleneck layer to interpret the model's focus, as shown in Figure \ref{fig:gradcam}.

The visualization confirms the efficacy of the proposed Parallel Context Block. The Left Panel displays the activation map of SpineContextResUNet, which demonstrates a cohesive and high-intensity focus strictly aligned with the vertebral bodies.

The constrained SwinUNETR \cite{he2023swinunetr} exhibits erroneous high-intensity activations on the ribs and soft tissue. Because the Transformer's capacity was artificially limited to fit the edge-hardware profile, its patch-based self-attention mechanism fails to properly converge on global spatial relationships, introducing blocky artifacts that confuse spinal structures with adjacent bone.
The Bottom Row illustrates the impact of our multi-dilated CNN design \cite{yu2015multi}, where increasing dilation rates progressively filters out these non-spinal artifacts, resulting in the clean localization seen in our final model. However, similar to the baseline 3D U-Net and ResUNet architectures, our model exhibits signal attenuation at the upper (cervical) and lower (sacral) regions. We attribute this to the inherent limitation of sliding-window inference, where the terminal patches lack bilateral context (padding effects), causing a drop in confidence at the volume boundaries relative to the center.

\section{Conclusion}
\label{sec:conclusion}

In this work, we presented SpineContextResUNet, a lightweight volumetric segmentation model designed to bridge the gap between high-performance AI and resource-constrained clinical environments. By using the Context Block, we successfully addressed the challenge of modeling long-range spinal dependencies without relying on computationally expensive mechanisms like Recurrent Neural Networks (RNNs) \cite{liao2018joint} or heavy Vision Transformers \cite{dosovitskiyimage}.

Our extensive evaluation on the CTSpine1K \cite{deng2025ctspine1k} and VerSe2020 \cite{sekuboyina2021verse} datasets demonstrates that SpineContextResUNet achieves robust binary segmentation performance (88.17\% Dice). Furthermore, we proved that under strict edge-hardware constraints, optimized CNN architectures with strong inductive biases vastly outperform heavily constrained Vision Transformers (SwinUNETR) in data-limited medical scenarios \cite{dosovitskiyimage}.

Crucially, we demonstrated that our model runs successfully on standard clerical hardware (Intel i5 CPU with 8GB RAM), whereas competing SOTA models unable to execute under memory constraints. This efficiency positions SpineContextResUNet as an ideal Stage 1 localizer for edge computing platforms like the Nvidia Jetson Orin Nano, enabling automated spinal analysis in remote or developing regions. Future work will focus on utilizing this rapid localization to crop volumes for fine-grained multi-class vertebrae identification, further reducing the computational barrier for comprehensive spinal diagnostics.

\section*{Acknowledgments}

The authors would like to thank the organizers and contributors of the VerSe2020 challenge \cite{sekuboyina2021verse} for providing the dataset used for training, validation, and internal testing of our proposed model. We also thank the creators of the CTSpine1K dataset \cite{deng2025ctspine1k}, which served as an essential resource for our external testing and generalization (domain-shift) evaluation. Furthermore, we thank the developers of the TotalSegmentator framework \cite{wasserthal2023totalsegmentator}, as their codebase was instrumental in establishing state-of-the-art baselines and conducting inference speed and memory usage comparisons.


\end{document}